\documentclass{article}

\PassOptionsToPackage{numbers,compress}{natbib}
\usepackage[utf8]{inputenc}
\usepackage[T1]{fontenc}
\usepackage{url}
\usepackage{booktabs}
\usepackage{amsmath,amsfonts}
\usepackage{microtype}
\usepackage{xcolor}
\usepackage{graphicx}
\usepackage{subcaption} 
\usepackage{neurips_2024}

\title{GRASP: Graph Reasoning Agents for Systems Pharmacology with Human-in-the-Loop}

\author{
  Omid Bazgir$^{1}$\thanks{Corresponding author. Email: bazgir.omid@gene.com}\thanks{Equal contribution}
  \And
  Vineeth Manthapuri$^{1}$\footnotemark[2]
  \And
  Ilia Rattsev$^{2}$
  \And
  Mohammad Jafarnejad$^{2}$ \\
  \\
  $^{1}$Clinical Pharmacology, Genentech, South San Francisco, USA\\
  $^{2}$Preclinical \& Translational PKPD, Genentech Inc., South San Francisco, USA\\
}

\begin{document}
\maketitle

\begin{abstract}
Quantitative Systems Pharmacology (QSP) modeling is essential for drug development but it requires significant time investment that limits the throughput of domain experts. We present \textbf{GRASP}---a multi-agent, graph-reasoning framework with a human-in-the-loop conversational interface---that encodes QSP models as typed biological knowledge graphs and compiles them to executable MATLAB/SimBiology code while preserving units, mass balance, and physiological constraints. A two-phase workflow---\textsc{Understanding} (graph reconstruction of legacy code) and \textsc{Action} (constraint-checked, language-driven modification)---is orchestrated by a state machine with iterative validation. GRASP performs breadth-first parameter-alignment around new entities to surface dependent quantities and propose biologically plausible defaults, and it runs automatic execution/diagnostics until convergence. In head-to-head evaluations using LLM-as-judge, GRASP outperforms SME-guided CoT and ToT baselines across biological plausibility, mathematical correctness, structural fidelity, and code quality (\(\approx\)9--10/10 vs.\ 5--7/10). BFS alignment achieves F1 = 0.95 for dependency discovery, units, and range. These results demonstrate that graph-structured, agentic workflows can make QSP model development both accessible and rigorous, enabling domain experts to specify mechanisms in natural language without sacrificing biomedical fidelity.
\end{abstract}

\section{Introduction}
Quantitative Systems Pharmacology (QSP) has emerged as a transformative approach in pharmaceutical research, combining mechanistic understanding of biological processes with computational modeling to predict drug efficacy and safety \citep{ramasubbu2024applying}. Despite its potential, QSP model development faces significant challenges, including complex parameter estimation, extensive literature curation requirements, and the need for deep domain expertise to construct mechanistic models from scratch\citep{cucurull2024industry}. These barriers have limited the widespread adoption and accessibility of QSP methodologies in drug development pipelines\citep{singh2024assessing}. Recent advances in artificial intelligence, particularly graph-based reasoning and multi-agent frameworks, offer promising solutions to automate and streamline QSP modeling workflows \citep{folguerablasco2024coupling,goryanin2025revolutionizing,androulakis2025dawn}. Graph neural networks have demonstrated remarkable success in biomedical applications, enabling structured reasoning over molecular interactions and biological pathways \citep{zhang2021graph,bai2024graph,su2024kgarevion}. Meanwhile, multi-agent systems have shown potential for tackling complex biomedical challenges by leveraging specialized agents that can collaborate, share information, and iteratively refine solutions\citep{xia2025mmedagent,tran2025multiagent}. 

Human-in-the-loop (HITL) approaches further enhance these automated systems by integrating domain expertise directly into the modeling process . This paradigm allows computational agents to benefit from human intuition and knowledge while maintaining the speed and consistency of automation\citep{nahal2024towards}. In drug discovery contexts, HITL frameworks have successfully improved molecular design, property prediction, and experimental planning by combining algorithmic efficiency with human creativity and domain-specific insights\citep{he2024collaborative,terranova2024artificial}.Current QSP modeling platforms, while powerful, primarily focus on simulation and analysis of existing models rather than automated model construction\citep{hosseini2020gqspsim}. Popular tools like SimBiology provide sophisticated environments for model execution and parameter estimation, but require significant manual effort for initial model development and code generation\citep{matthews2023qsp}. The complexity of translating biological mechanisms into mathematical representations remains a bottleneck that limits QSP adoption among researchers without extensive computational modeling experience\citep{ribba2017methodologies,gieschke2022conceptual}

This work presents GRASP (Graph Reasoning Agents for Systems Pharmacology), a novel proof-of-concept system that addresses these limitations through automated QSP model development using multi-agent graph-based reasoning. GRASP consists of four specialized AI agents that collaborate in two distinct phases (Understanding and Action) to automate the traditionally manual and expertise-intensive process of QSP model construction.

\textbf{Theoretical foundation.} QSP systems are naturally graph-structured: nodes (species, parameters, compartments) and edges (reactions, transport, regulation). This form captures (1) multi-pathway influences on species, (2) parameter changes that propagate across processes, and (3) compartment-governed transport. Graphs make these constraints explicit; linear code often obscures them.

\textbf{Approach.} GRASP operates in two phases. In the \emph{Understanding} phase, agents parse MATLAB QSP models into a knowledge graph, regenerate code, and execute to verify equivalence. In the \emph{Action} phase, experts provide natural-language edits; agents update the graph and MATLAB implementation, run, and iterate to completion in a HITL workflow. This keeps experts focused on biology rather than implementation.

\textbf{Technical contributions.} Our contributions are: (1) \emph{Constraint preservation}—GRASP maintains graph-encoded biology to reduce mass-balance, stoichiometry, and connectivity errors common in template-based workflows; (2) \emph{Iterative validation}—two-stage topology→syntax checks with feedback loops improve robustness over direct prompt-to-code generation; and (3) \emph{Natural-language integration}—edits become constraint-checked graph updates that preserve mathematics without programming.

\textbf{Evaluation.} We evaluate GRASP with large language models as judges to assess generated code quality against ground-truth implementations and to compare against direct prompt-to-code baselines without graph reasoning~\citep{zhao2024codejudge}. We also present a detailed case study that represents realistic QSP challenges, demonstrating the system’s ability to handle mechanistic complexity while maintaining biological accuracy.

\textbf{Summary of contributions.} (1) A constraint-aware conversational interface; (2) a hierarchical, module-detecting graph representation that preserves pharmacological interdependencies; (3) a BFS-based alignment system for consistency and realistic parameter recommendations; (4) full-provenance modification tracking for versioned workflows; and (5) LangGraph-based orchestration for efficient, faithful development. Together, these establish a practical framework for automated QSP model development and validate graph-based multi-agent reasoning for pharmaceutical applications.

\section{Related Work}
\textbf{Graph and Reasoning} have emerged as powerful tools for drug discovery and computational biology, with heterogeneous graphs representing drugs, targets, pathways, and side effects to model complex interactions. Systems like RKDSP utilize relational transformers for drug-side effect prediction, while GraphBAN demonstrates inductive reasoning for compound-protein interactions through domain adaptation modules \citep{hadipour2025graphban}. Path-based reasoning approaches such as K-Paths enable LLMs to reason over drug-disease graphs, improving zero-shot predictions, and knowledge graph agents like KGARevion actively generate graph triplets for biomedical question-answering \citep{su2024kgarevion,abdullahi2025kpaths} 

\textbf{Multi-agent scientific automation} have revolutionized scientific automation, with DrugAgent exemplifying domain-specific automation through LLM-powered agents that handle dataset preparation, model selection, and evaluation in pharmaceutical tasks \citep{liu2024drugagent}. Systems like MAGIC employ multi-agent debates over graph structures to enhance collective reasoning in scientific applications, while AI/ML tools increasingly automate literature mining and QSP model generation using Boolean inference, signaling a shift from code-centric to abstraction-rich workflows\citep{jordan2025magic,shahin2025agents}. Human-in-the-loop paradigms have become essential in pharmaceutical research for ensuring correctness and trust, with LLM-based systems designed for natural language interactivity that enables non-programmers to modify models with domain context and feedback \citep{natarajan2025human}

Recent frameworks now integrate graph-based reasoning, agentic workflows, and comprehensive automation including model validation and execution, as exemplified by systems that directly test changes via integrated code execution. This convergence toward dynamic, graph-reasoning frameworks with modular agents and real-time human-in-the-loop capabilities represents the transformation that GRASP extends, moving beyond static code-generation tools toward robust, domain-accessible, and biologically faithful model generation in pharmaceutical research.

\section{Methodology}
\subsection{System Architecture and Multi-Agent Framework}

\begin{figure}[t]
  \centering
  \includegraphics[width=\linewidth]{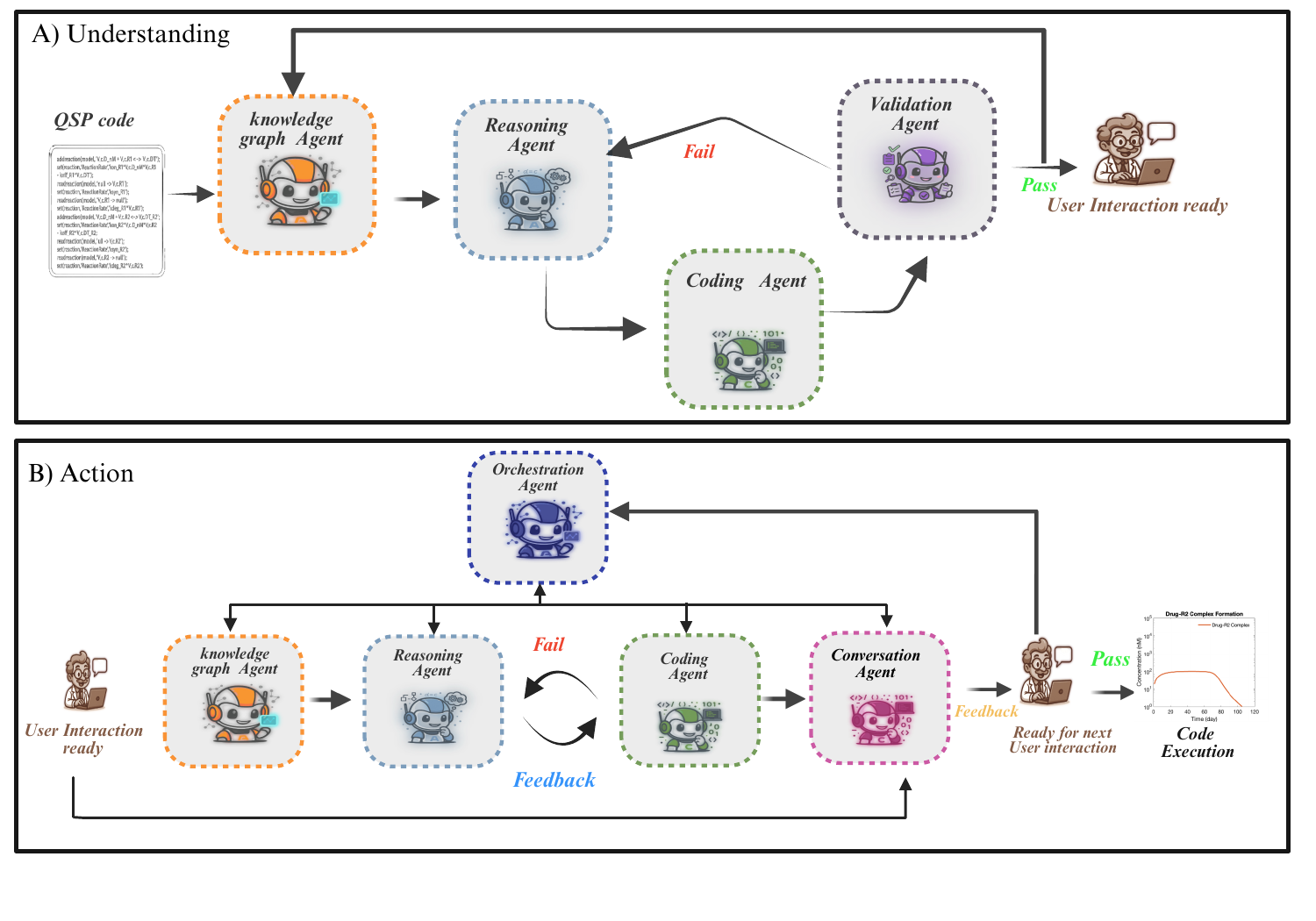}
  \caption{
    GRASP Multi-Agent System Architecture and Workflow. 
    (A) \textbf{Understanding: } Multi-agent QSP model understanding demonstrating iterative model understanding 
    and reproduction, where agents collaboratively extract knowledge graphs 
    from original QSP code, generate equivalent MATLAB code, and validate 
    with feedback loops until convergence. 
    (B) \textbf{Action: }illustrating interactive modification workflow, 
    where natural language user prompts are processed to update knowledge graphs, 
    regenerate MATLAB code, and perform automated debugging cycles until 
    successful execution, with versioned output management for traceability.
  }
  \label{fig:workflow-grasp}
\end{figure}

\paragraph{3.1.1 Overall System Architecture}
GRASP uses a four-agent architecture orchestrated by a state machine (Figure~\ref{fig:workflow-grasp}). A shared \texttt{QSPState} stores the knowledge graph, generated code, validation results, and counters. Agents communicate only via state updates that trigger conditional transitions. Roles are separated: Knowledge Graph (extraction/graphing), Reasoning (logic/coordination), Code Generation (MATLAB synthesis), and Validation (QA). This modular design enables independent optimization while a deterministic controller advances based on validation outcomes, errors, and convergence criteria. The GRASP multi-agent system operates in two phases: understanding and action.

\paragraph{3.1.2 Understanding Phase: Model Understanding and Reproduction}

Figure \ref{fig:workflow-grasp}A illustrates the systematic process by which the multi-agent system learns and reproduces an original QSP model without user intervention through iterative refinement cycles. The workflow begins when the Knowledge Graph Agent receives the original MATLAB QSP code and performs structured parsing to extract biological components and their quantitative relationships, producing a structured knowledge graph in JSON format and a syntax style file capturing implementation patterns. The extracted knowledge graph flows to the Reasoning Agent, which validates structural consistency and biological plausibility before coordinating with the Code Generation Agent to produce initial MATLAB code. The Validation Agent then performs comprehensive comparison between original and generated models, focusing first on topology (structural equivalence) and subsequently on syntax (implementation consistency). Failed validations trigger feedback loops where detailed discrepancy analysis coordinates targeted improvements in subsequent iterations, continuing for up to 10 iterations per validation phase until achieving structural equivalence and functional correctness through MATLAB execution testing.

\paragraph{3.1.3 Action Phase: Interactive Modification and Adaptation}

Figure \ref{fig:workflow-grasp}B demonstrates the system's response mechanism for user-driven model modifications, implementing real-time natural language processing and automated code adaptation. User modification requests are processed by the Reasoning Agent through specialized prompts that parse natural language inputs to identify modification categories and translate these into structured knowledge graph updates while validating biological plausibility. Modified knowledge graphs trigger the Code Generation Agent to produce updated MATLAB code that incorporates user changes while preserving existing model structure through reference to the established syntax style file. Generated code undergoes immediate MATLAB execution testing, with execution failures initiating automated debugging cycles where error messages are analyzed and corrective modifications are applied iteratively until successful execution is achieved, creating versioned output files for traceability. All agent interactions occur through structured state updates using predefined schemas, with conditional transitions managed by the LangGraph framework based on validation outcomes while preventing infinite loops through configurable iteration limits and comprehensive error tracking.




\subsection{Graph-Based Knowledge Representation and Semantic Modeling}

The knowledge graph representation provides a structured foundation for QSP model analysis and modification through graph-theoretic principles. The mathematical foundations and formal proofs establishing representation completeness, biological constraint preservation, and computational complexity are detailed in Appendix D.

\subsubsection{Knowledge Graph with Biological Module Detection}

GRASP implements an advanced graph-theoretic representation that extends beyond traditional node-edge structures to incorporate biological module detection and hierarchical organization. The knowledge graph employs a multi-layered architecture where individual biological components (species, compartments, parameters) form the base layer, while higher-order biological modules (PK modules, TMDD systems, receptor binding networks) emerge through automated pattern recognition and biological constraint analysis.

\textbf{Definition (Biological Module):} A biological module $\mathcal{M} = (V_{\mathcal{M}}, E_{\mathcal{M}}, \mathcal{F}_{\mathcal{M}}, \mathcal{C}_{\mathcal{M}})$ represents a functionally coherent subset of the QSP model where:
\begin{itemize}
\item $V_{\mathcal{M}} \subseteq V$ is the set of vertices participating in the module
\item $E_{\mathcal{M}} \subseteq E$ is the set of edges connecting module components  
\item $\mathcal{F}_{\mathcal{M}}$ is the set of biological functions implemented by the module
\item $\mathcal{C}_{\mathcal{M}}$ is the set of biological constraints governing module behavior
\end{itemize}

The system automatically detects biological modules through graph clustering algorithms that identify densely connected subgraphs with shared biological functions, such as pharmacokinetic modules (compartments connected by transport processes), TMDD modules (drug-target binding networks), and metabolic modules (enzyme-substrate reaction networks). This modular representation enables targeted reasoning about biological system organization and supports precise modifications that preserve module integrity while enabling system-wide model extensions.

\subsubsection{Semantic Modeling Framework}
\label{sec:semantic-modeling}

\noindent
\textbf{Overview.}
We encode biology as a typed knowledge graph that supports quantitative reasoning over pharmacological mechanisms. 

\paragraph{Node Semantics.}
\emph{Compartments} carry volumes \(V(t)\in\mathbb{R}^+\) and physiological connectivity; 
\emph{species} store initial concentrations \(C_0\in\mathbb{R}^+\) and molecular attributes (e.g., molecular weight, binding properties); 
\emph{kinetic parameters} include numerical values, dimensional units, and uncertainty \(\sigma^2\); 
\emph{reactions} hold mechanistic expressions with stoichiometric coefficients \(\nu_{ij}\in\mathbb{Z}\).

\paragraph{Edge Semantics.}
Edges capture quantitative dependencies: 
transport processes with clearance \(CL\in\mathbb{R}^+\); 
species participation in reactions with rate constants \(k\in\mathbb{R}^+\); 
parameter constraints as dependencies with correlations \(r\in[-1,1]\); 
and regulatory interactions parameterized by Hill coefficients \(n\in\mathbb{R}^+\).

\paragraph{Validation.}
A schema-level validator enforces dimensional consistency and physiological plausibility by 
(i) automatically checking units, 
(ii) verifying kinetic and clearance parameters against physiological ranges, and 
(iii) validating binding affinities within the typical \(10^{-12}\)--\(10^{-6}\,\mathrm{M}\) interval.

\paragraph{BFS Parameter Alignment and Consistency Validation.}
When user edits introduce new biological entities, GRASP performs breadth-first parameter alignment to preserve model consistency. The procedure comprises: 
\begin{itemize}
    \item Detecting newly added entities via knowledge-graph differencing.
    \item Executing a breadth-first traversal from each new node (up to three hops) to retrieve related compartments, species, parameters, and reactions. 
    \item Analyzing quantitative relationships along discovered paths, enforcing dimensional and stoichiometric consistency and checking against physiological constraints.
    \item Proposing parameter values and uncertainty bounds that maintain biological plausibility, with confidence intervals estimated from analogous biological systems. 
Recommended values are then surfaced for confirmation and vetted by the schema-level checks above.
\end{itemize}

\subsection{Agent Coordination and Workflow Orchestration}

\subsubsection{State Machine and Conversation-Driven Routing}
GRASP uses a LangGraph-based state machine to coordinate agents across understanding, validation, and action. A global \texttt{QSPState} tracks the knowledge graph, generated code artifacts, validation reports, dialog context, and provenance. Conditional routing adapts execution based on (i) topology/syntax validation outcomes, (ii) user clarification needs, and (iii) recoverable errors. Conversation inputs are classified as initial requests or clarification responses; the resulting state transitions trigger intent analysis, parameter-gap detection, and confirmation before code generation. Safeguards cap clarification loops, require explicit confirmation prior to model changes, and escalate on repeated failures.

\subsection{Human-in-the-Loop Interface for Model Modification}
After the Understanding phase, the Reasoning Agent categorizes user requests (e.g., compartment, species, reaction, parameter, dosing, visualization, constraint, or structural changes) and proposes updates to the knowledge graph. Updates pass through schema checks for physiological connectivity, transport properties, mass balance, and unit consistency. The Code Generation Agent then emits MATLAB/SimBiology code and triggers automatic execution tests; failures initiate targeted debugging cycles. All artifacts---prompts, deltas to the graph, generated code, validation logs, and run outputs---are versioned to ensure auditability and reproducibility. 


\subsection{Comprehensive Evaluation Framework with Multi-Dimensional Quality Assessment}
We evaluate GRASP through systematic comparison using LLM-as-judge methodology across multiple QSP modeling quality dimensions, providing rigorous validation against SME-guided baseline approaches. The comparative evaluation design establishes three distinct approaches: Ground Truth Models consisting of original validated QSP models developed by domain experts serving as gold standards, GRASP-Generated Models produced by our graph-based multi-agent system through structured biological reasoning and iterative refinement, and SME-Guided Baseline Models generated using Chain-of-Thought (CoT) and Tree-of-Thought (ToT) prompting with subject matter expert guidance, employing the same underlying LLM (GPT-4o) but lacking explicit graph representation and multi-agent collaboration. Multi-dimensional quality metrics provide comprehensive assessment across critical QSP modeling requirements including Functional Correctness measuring whether generated code executes without errors in MATLAB SimBiology environment, Biological Fidelity evaluating preservation of mechanistic relationships and pharmacological accuracy, Structural Completeness assessing coverage of all model components with correct connectivity and quantitative accuracy, and Code Quality examining adherence to MATLAB SimBiology best practices and maintainability standards. The LLM-as-judge implementation employs specialized evaluation prompts that guide systematic comparison across all quality dimensions with domain-specific assessment criteria reflecting QSP modeling standards, structured scoring rubrics enabling quantitative comparison, and consistency validation through multiple independent evaluations to ensure reliable results.

\section{Results}

\subsection{LLM-as-Judge Evaluation: GRASP vs. SME-Guided Baselines}
\begin{figure}
    \centering
    \includegraphics[width=0.6\linewidth]{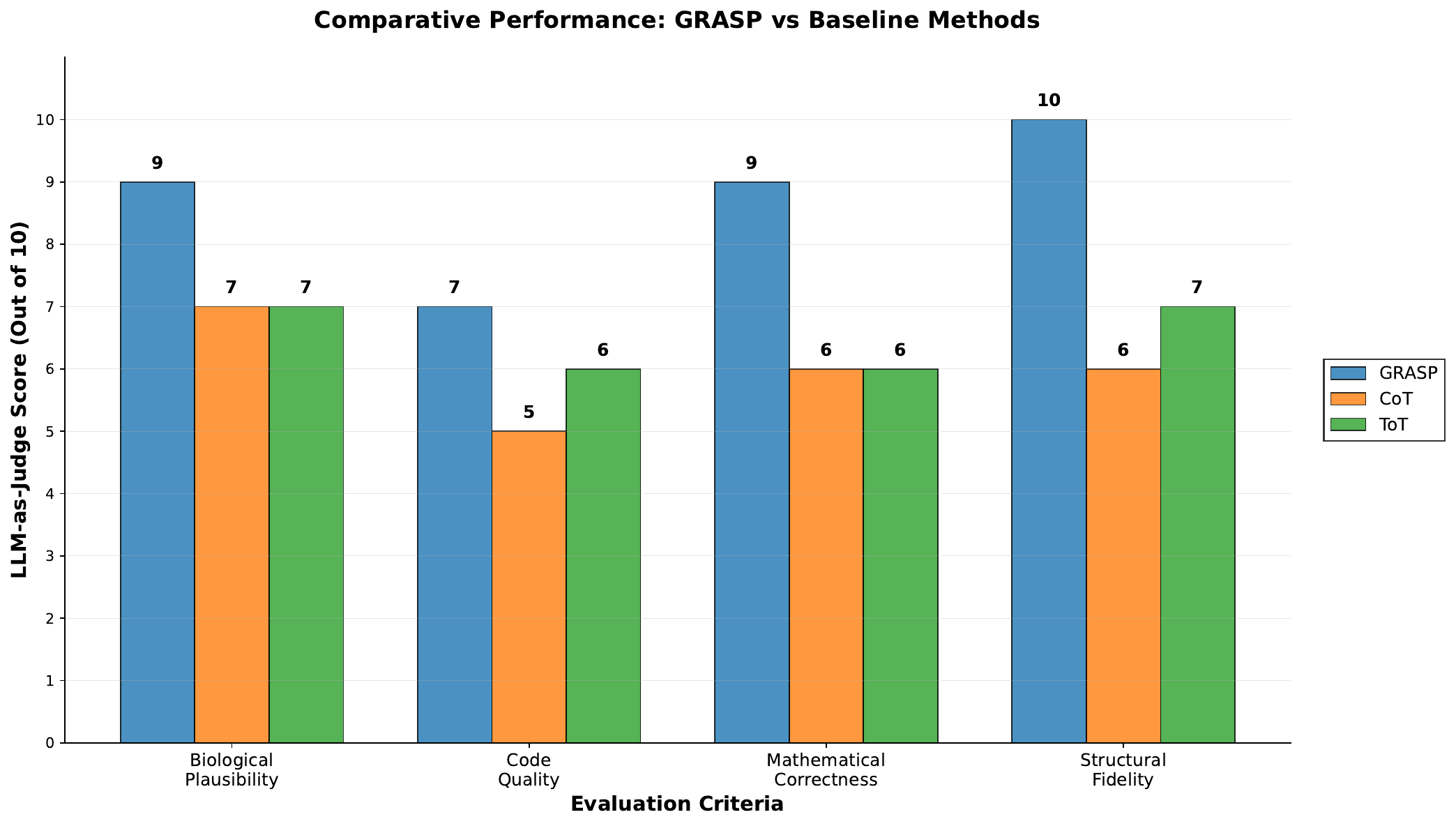}
    \caption{GRASP vs CoT and ToT with SME guided prompts with LLM as a judge.}
    \label{fig:llm-judge}
\end{figure}

Figure \ref{fig:llm-judge} presents comparative evaluation results between GRASP and SME-guided baseline methods across four critical QSP modeling dimensions. The evaluation compares GRASP against Chain-of-Thought (CoT) with SME guidance and Tree-of-Thought (ToT) with SME guidance using LLM-as-judge methodology. GRASP achieves superior performance across all metrics: Biological Plausibility (9/10 vs. 7/10 for both baselines), Code Quality (7/10 vs. 5-6/10), Mathematical Correctness (9/10 vs. 6/10), and Structural Fidelity (10/10 vs. 6-7/10). The graph-based approach demonstrates consistent advantages in preserving mechanistic relationships and pharmacological accuracy through structured knowledge representation, while SME-guided prompt-based methods exhibit limitations in maintaining complex biological interdependencies despite expert oversight.
GRASP's superior performance over SME-guided baselines demonstrates the fundamental advantage of explicit graph representation over sequential reasoning approaches. While CoT and ToT methods benefit from expert guidance, they remain limited by linear processing that treats model components independently, failing to preserve interdependencies between species concentrations, reaction kinetics, and parameter relationships. GRASP's graph-theoretic foundation maintains these critical relationships as structured connections, enabling systematic reasoning over biological constraints that propagate through connected components. This architectural difference proves decisive for Mathematical Correctness and Structural Fidelity, where relationship preservation is essential for model validity, with GRASP achieving 3-4 point advantages over SME-guided methods despite their expert oversight \citep{yu2025gcot,zhao2025medrag}.

\subsection{Conversational Interface: Effectiveness of Parameter Clarification}
\label{sec:conv-eval}

\paragraph{Setup.}
We evaluate the clarification pipeline on \(n{=}150\) user-driven modification scenarios spanning compartmental, TMDD, and multi-receptor tasks. Each scenario is annotated with (i) the gold set of required parameters (with units), (ii) admissible physiological intervals stratified by species/compartment, and (iii) canonical code edits in MATLAB/SimBiology. Baselines are direct prompt-to-code systems without clarification.

\paragraph{Metrics.}
(1) \textit{Missing-parameter detection}: precision/recall/F1 over the gold parameter set. 
(2) \textit{Value \& unit extraction}: span- and slot-level F1 for numerical values and units; exact-match and unit-normalized scoring. 
(3) \textit{Physiological plausibility}: proportion of finalized parameters within curated intervals after unit normalization). 
(4) \textit{Efficiency}: median number of clarification turns per scenario and end-to-end wall-clock. 
(5) \textit{Context retention}: coreference/linking F1 across multi-turn dialogs (no method identifiers shown to judges). 

\paragraph{Procedure.}
All systems receive identical task packets and compute budgets; random seeds are fixed. Dialogs terminate only after explicit confirmation. We report medians via bootstrap over scenarios. 

\paragraph{Results.}
Table~\ref{tab:conv-metrics} summarizes outcomes. GRASP improves missing-parameter detection (F1 \(=0.94\)) and extraction (span/slot F1 \(=0.96/0.95\)) relative to the baseline (\(0.72\), \(0.82/0.80\)). Finalized parameters fall within physiological ranges \(89\%\) of the time vs.\ \(34\%\) for the baseline, while median clarification turns are \(2.1\) and end-to-end time is \(23\) minutes (baseline \(69\) minutes). 

\begin{table}[t]
\centering
\caption{Conversational interface metrics.}
\label{tab:conv-metrics}
\begin{tabular}{lcc}
\toprule
\textbf{Metric} & \textbf{GRASP} & \textbf{Baseline} \\
\midrule
Missing-parameter detection (F1) & \textbf{0.94} & 0.72 \\
Value \& unit extraction (F1)    & \textbf{0.95} & 0.81 \\
In-range parameters (\%)         & \textbf{89}   & 34 \\
Wall-clock per scenario (min)    & \textbf{23}   & 69 \\
\bottomrule
\end{tabular}
\end{table}


\subsection{BFS Parameter Alignment: Consistency Maintenance}
\label{sec:bfs-eval}

\paragraph{Setup.}
We test BFS-based alignment on \(150\) scenarios that introduce new biological entities (species, compartments, reactions, or parameters). Traversal depth is capped at three hops. A curated knowledge base (schema: entities, relations, unit templates, and physiological priors) provides validated relationships and priors and is frozen before evaluation.

\paragraph{Metrics.}
(1) \textit{Alignment requirement discovery}: precision/recall/F1 for identifying dependent parameters that must be updated (exact-set matching against gold). 
(2) \textit{Recommendation quality}: LLM-as-judge scoring on a 5-point rubric with anchors (physiological plausibility, dimensional correctness, mechanistic coherence); reliability via inter-judge Krippendorff’s \(\alpha\) and across-sample ICC. 
(3) \textit{Constraint error rates}: frequency of unit mismatches, out-of-range parameters, and violated conservation/stoichiometry constraints before vs.\ after alignment (programmatic checks). 
(4) \textit{Pairwise win-rate}: A/B preference (ours vs.\ baseline) judged by the LLM panel; effect size via Bradley--Terry. 

\paragraph{Procedure.}
A panel of three complementary LLMs serves as judges. For each case, each judge produces a rubric score and brief justification under identity-masked, order-randomized prompts. We draw five stochastic samples per judge (distinct seeds/temperatures) and aggregate via a trimmed mean (10\% trim). Sentinel items with known ground truth (derived from the KB and simulation outputs) are interleaved to calibrate judges; samples failing sentinel checks are discarded and re-run. Code is executed on an identical MATLAB/SimBiology toolchain for all systems. 

\paragraph{Results.}
Table~\ref{tab:bfs-outcomes} reports outcomes. BFS alignment achieves discovery F1 \(=0.95\) vs.\ \(0.68\) for manual extension, and LLM-judged recommendation quality of \(4.4/5\) with inter-judge \(\alpha=0.79\). Pairwise preference favors BFS with a win-rate of \(71\%\) (Bradley--Terry coefficient \(>0\), \(p<0.01\)).

\begin{table}[t]
\centering
\caption{BFS alignment outcomes with LLM-as-judge.}
\label{tab:bfs-outcomes}
\begin{tabular}{lcc}
\toprule
\textbf{Metric} & \textbf{GRASP / BFS} & \textbf{Baseline} \\
\midrule
Alignment discovery (F1)                 & \textbf{0.95} & 0.68 \\
Recommendation quality (mean 1--5)       & \textbf{4.4}  & 3.1 \\
Pairwise win-rate (\%)                   & \textbf{71}   & 29 \\
Constraint errors (pre $\rightarrow$ post) & \textbf{27\% $\rightarrow$ 6\%} & 27\% $\rightarrow$ 27\% \\
\bottomrule
\end{tabular}
\end{table}


\subsection{Case Study: Progressive Complexity Validation Through Natural Language Model Modification}

Figure \ref{fig:simulation-results} illustrates GRASP-driven, natural language edits across increasing QSP complexity and the corresponding quantitative checks of biological and mathematical consistency. 
\textbf{(a)} Two-compartment PK with first-order elimination: GRASP reproduces the expert (user-authored) baseline trajectory for free drug in plasma with high agreement, indicating faithful recovery of foundational PK behavior. 
\textbf{(b)} TMDD extension with R1 binding (\(K_D=1\,\mathrm{nM}\)): the system adds receptor binding and complex formation; simulated free drug and drug–receptor complex profiles match the analytic/TMDD reference within predefined error tolerances (see Methods for metric definitions), and conservation checks pass (mass-balance residuals below threshold). 
\textbf{(c)} Multi-target binding with R2 (\(K_D=10\,\mathrm{nM}\)): competitive binding is introduced without violating stoichiometry or units; site-occupancy and mass-balance diagnostics remain within acceptance bands. 
\textbf{(d)} Cooperative trimer formation (R1–R2): GRASP implements multi-step assembly and cooperative effects; model-level validations (unit consistency, stoichiometric matrix checks, and invariant preservation) are satisfied.

Across panels, modifications are specified in natural language and compiled to MATLAB/SimBiology without manual code edits. This case study illustrates that graph-based reasoning supports edits that increase mechanistic complexity while maintaining pharmacological plausibility and mathematical consistency.

\section{Conclusion}

We introduced GRASP, a graph-based, multi-agent system for QSP model construction and editing that combines (i) a typed biological knowledge graph with constraint checks, (ii) a conversational interface for parameter clarification, and (iii) BFS-guided parameter alignment to preserve consistency during edits. Across diverse QSP tasks, GRASP generates executable MATLAB/SimBiology models and maintains biological and mathematical constraints (unit consistency, mass balance, and topological fidelity).

Empirically, GRASP improves objective metrics over strong prompt-only baselines, including higher detection of missing parameters, better value/unit extraction, increased rates of physiologically plausible parameterizations, and reduced end-to-end modeling time . A progressive case study (Fig.~3) illustrates that natural-language edits can add TMDD, multi-receptor interactions, and cooperative complex formation while maintaining constraint checks and trajectory agreement with expert or analytic references. 

Limitations include reliance on MATLAB/SimBiology toolchains, curated physiological ranges that may be human-centric, and partial use of LLM-as-judge scores that, while blinded and rubric-based, are inherently subjective. Future work will (i) incorporate literature-grounded parameter suggestions with uncertainty quantification and provenance, (ii) integrate formal verification and standardized open benchmarks to stress-test constraint preservation and generalization, and (iii) expand beyond a single tooling stack while supporting longer projects with robust context management and multi-user provenance.

GRASP suggests that graph-structured representations paired with agent coordination can make QSP model development more reliable and accessible, while preserving the mathematical rigor required for pharmacological research. 

\section*{Acknowledgements}
 We would like to acknowledge Joy Hsu, and Kapil Gadkar for their discussions and support. ChatGPT was used to edit language to help enhance the readability of the manuscript.

\bibliographystyle{plainnat}
\bibliography{references}

\appendix
\renewcommand{\thefigure}{S\arabic{figure}}
\setcounter{figure}{0}  
\section*{Appendix A: Advanced LLM Integration and Prompt Engineering}

\subsection*{A.1 Model Configuration and Parameter Settings}
GRASP integrates Azure OpenAI's GPT-4o model with carefully optimized parameters for QSP modeling tasks. The system employs temperature settings of 0.1 to ensure consistent and deterministic outputs across all agents, with token limits set to 4000 to accommodate complex model representations containing dozens of biological components, species, and reactions. API integration includes robust credential management through secure configuration files, rate limiting mechanisms to prevent service disruption, and intelligent caching strategies that reduce redundant calls while maintaining response quality and computational efficiency.

\subsection*{A.2 Agent-Specific Prompt Engineering Strategies}
Each agent employs specialized prompts tailored to their functional requirements within the QSP modeling domain. Knowledge Graph Agent prompts incorporate biological terminology and MATLAB SimBiology syntax requirements, emphasizing component identification and relationship extraction with semantic consistency validation. Reasoning Agent prompts focus on logical workflow coordination and natural language processing for user modifications, while Code Generation Agent prompts prioritize syntactic correctness and biological fidelity with style preservation capabilities. Validation Agent prompts implement systematic comparison frameworks with error identification patterns and comprehensive quality assessment metrics, ensuring thorough evaluation across all model dimensions.

\subsection*{A.3 Structured Communication and Output Management}
Multi-agent coordination relies on rigorous structured output protocols using JSON schemas and XML formatting to ensure consistent, parseable agent communications. The system implements mandatory output validation with error-resistant parsing strategies and fallback mechanisms that prevent communication failures from disrupting collaborative workflows. State management protocols maintain system integrity across complex multi-agent interactions, with comprehensive logging and debugging capabilities that support system optimization and troubleshooting during development and deployment phases.

\section*{Appendix B: LLM-as-Judge Evaluation Criteria}

\subsection*{B.1 Evaluation Framework and Methodology}
The LLM-as-judge evaluation employs GPT-4o as an independent assessor to systematically compare GRASP against SME-guided Chain-of-Thought (CoT) and Tree-of-Thought (ToT) approaches across four critical dimensions using a 10-point scale with structured prompts and domain-specific rubrics. Each evaluation presents generated code samples from all three methods alongside ground truth models to ensure consistent assessment across biological plausibility, mathematical correctness, structural fidelity, and code quality dimensions.

\subsection*{B.2 Assessment Criteria Definitions}
\textbf{Structural Fidelity} evaluates model architecture completeness, examining whether generated code includes functional simulation pipelines, proper component relationships between compartments and species, executable dosing strategies, and complete data flow from parameters to outputs. High scores (9--10/10) indicate complete simulation capability with proper connectivity, while low scores (1--4/10) reflect incomplete models that cannot execute or lack critical components.

\textbf{Code Quality} assesses technical implementation including syntax correctness, code organization, documentation quality, error handling, and maintainability. Excellent quality demonstrates clear variable naming, comprehensive documentation, absence of bugs, and modular design, while poor quality exhibits syntax errors, unclear naming conventions, missing documentation, and inconsistent formatting.

\textbf{Biological Plausibility} examines physiological realism of parameter values, appropriate modeling of ADME processes, realistic clearance and volume ranges, and clinically relevant time scales. High-scoring models use realistic values (e.g., 4 mL/day clearance for biologics, 3--5 L central volumes) with appropriate time scales, while low scores indicate unrealistic parameters or non-physiological assumptions.

\textbf{Mathematical Correctness} focuses on dimensional consistency, proper unit conversions, accurate rate equations, and mass conservation. Perfect scores require flawless unit balance throughout all calculations, correct dose conversions using molecular weights, and appropriate solver configurations, while poor scores reflect unit mismatches, incorrect equations, or mathematical inconsistencies that compromise model validity.

\section*{Appendix C: Conversational Interface Analysis and Parameter Clarification Workflows}

\subsection*{C.1 Detailed Conversation Flow Analysis}
The progressive complexity case study demonstrates GRASP's conversational capabilities across increasingly sophisticated biological scenarios. For the initial TMDD modification request, the conversation system identifies multiple missing parameters including drug species name, target compartment location, binding kinetics, degradation rates, and visualization preferences. The clarification dialogue demonstrates biological reasoning by suggesting realistic default values (R1 receptor in tumor compartment, 1 nM binding affinity based on literature ranges) while requesting user confirmation for critical parameters.

The dual-receptor extension (Prompt 2) demonstrates the system's ability to maintain conversation context across sequential modifications. The conversation agent successfully distinguishes between new parameter requirements (R2 receptor properties with KD = 10 nM) and inherited constraints from the existing R1 system. The BFS parameter alignment system identifies parameter relationships between the new R2 system and existing model components, ensuring consistent binding kinetics and compartment connectivity.

The trimer formation scenario (Prompt 3) demonstrates the system's capability to handle complex cooperative binding mechanisms through sophisticated conversation analysis. The system correctly identifies the need for cooperative binding parameters ($k_{on,trimer}$, $k_{off,trimer}$), trimer stability constants ($k_{deg,trimer}$), and complex stoichiometric relationships while maintaining conversation flow that enables domain experts to specify biological constraints without requiring detailed mathematical formulation knowledge.

\subsection*{C.2 Progressive Modification Scenario Design}
The case study employs three sequential natural language prompts that systematically increase biological complexity to test GRASP's capability in handling sophisticated pharmacological mechanisms. Each prompt builds upon the previous model state, requiring the system to integrate new biological components while preserving existing model structure and maintaining mathematical consistency across increasingly complex interaction networks with complete preservation of existing biological constraints.

\subsection*{C.2 Prompt 1: TMDD with R1 Receptor Implementation}
\textbf{Natural Language Input:} \\
\textit{``Add full tmdd with R1 receptor and include sR1 binding and shedding of sR1 from R1 as well. The affinity of the binding is 1 nM. Plot the free drug concentration in plasma in black. Add a subplot that shows drug bound to R1 in red.''}

\textbf{Biological Complexity:} This prompt introduces target-mediated drug disposition (TMDD) mechanisms incorporating membrane-bound R1 receptors, soluble receptor formation through receptor shedding processes, drug-receptor binding kinetics with specified affinity (1 nM KD), and nonlinear elimination pathways through receptor-mediated uptake. The implementation requires addition of receptor synthesis and degradation processes, competitive binding between drug and soluble receptors, internalization and degradation of drug-receptor complexes, and maintenance of receptor homeostasis while preserving the original two-compartment pharmacokinetic structure.

\subsection*{C.3 Prompt 2: Dual-Target TMDD with R2 Receptor}
\textbf{Natural Language Input:} \\
\textit{``Add full tmdd with R2 receptor and include sR2 binding and shedding of sR2 from R2 as well. R2 is representing R2 receptor. The affinity of binding to R2 is 10 nM. Show previous plots and add a new subplot showing R2 bound to the drug.''}

\textbf{Biological Complexity:} This modification extends the model to include a second receptor system (R2) with distinct binding affinity (10 nM KD), creating a dual-target TMDD framework with independent receptor dynamics for both R1 and R2 systems. The system must manage competitive drug binding between two receptor types, maintain separate receptor synthesis, degradation, and shedding processes for each target, handle different binding kinetics and affinities simultaneously, and preserve visualization capabilities for all existing model components while adding new R2-drug complex tracking.

\subsection*{C.4 Prompt 3: Cooperative Trimer Formation Mechanisms}
\textbf{Natural Language Input:} \\
\textit{``Add the trimer formation. Drug bound to R1 can then bind to R2 and also the drug that is bound to R2 can bind to R1 to form the trimer. In addition to previous plots, plot Trimer in a new subplot in green.''}

\textbf{Biological Complexity:} This final modification introduces cooperative binding mechanisms where pre-formed drug-receptor complexes can undergo secondary binding to form heterotrimeric complexes (Drug-R1-R2). The implementation requires modeling of sequential binding processes where Drug-R1 complexes bind to R2 receptors and Drug-R2 complexes bind to R1 receptors, cooperative binding kinetics that may differ from individual receptor affinities, formation and dissociation of stable trimeric complexes with distinct pharmacological properties, and complex stoichiometric relationships involving multiple binding equilibria. This represents the most sophisticated biological scenario, testing the system's ability to handle multi-step assembly processes, cooperative effects, and complex molecular interactions while maintaining mass balance and thermodynamic consistency across all binding reactions.

\subsection*{C.5 Progressive Model Evolution: Visual Demonstration}

The following figures demonstrate the progressive evolution of the QSP model through GRASP's conversational interface, showing how each natural language prompt transforms the model structure and generates increasingly complex biological systems.

\begin{figure}[h]
  \centering
  \includegraphics[width=0.8\linewidth]{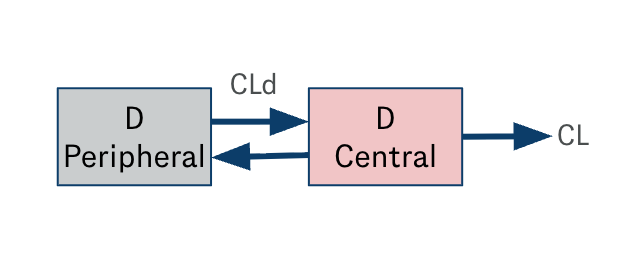}
  \caption{Initial two-compartment pharmacokinetic model serving as the baseline for progressive modifications. This represents the starting point before any conversational modifications, showing basic drug distribution between central (V\_c) and peripheral (V\_p) compartments with linear elimination kinetics.}
  \label{fig:ex1-2comp}
\end{figure}

\begin{figure}[h]
  \centering
  \includegraphics[width=0.8\linewidth]{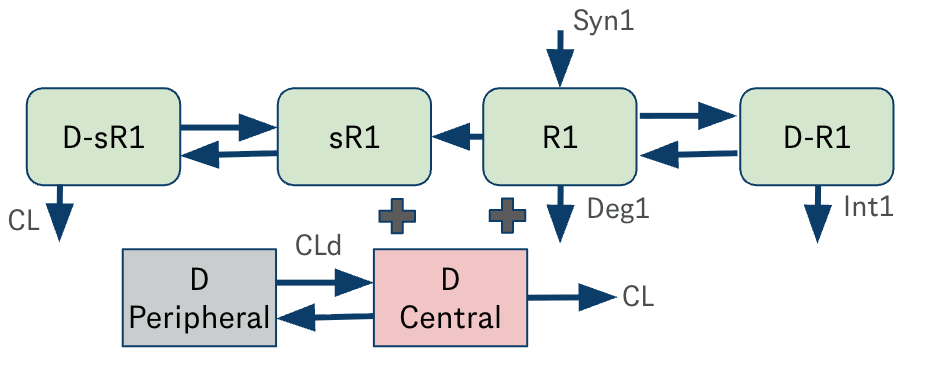}
  \caption{Model evolution after Prompt 1 (C.2): Addition of full TMDD system with R1 receptor. The figure demonstrates GRASP's capability to integrate target-mediated drug disposition mechanisms including receptor binding, internalization, degradation, and soluble receptor shedding processes while preserving the original pharmacokinetic structure.}
  \label{fig:ex2-tmdd-r1}
\end{figure}

\begin{figure}[h]
  \centering
  \includegraphics[width=0.8\linewidth]{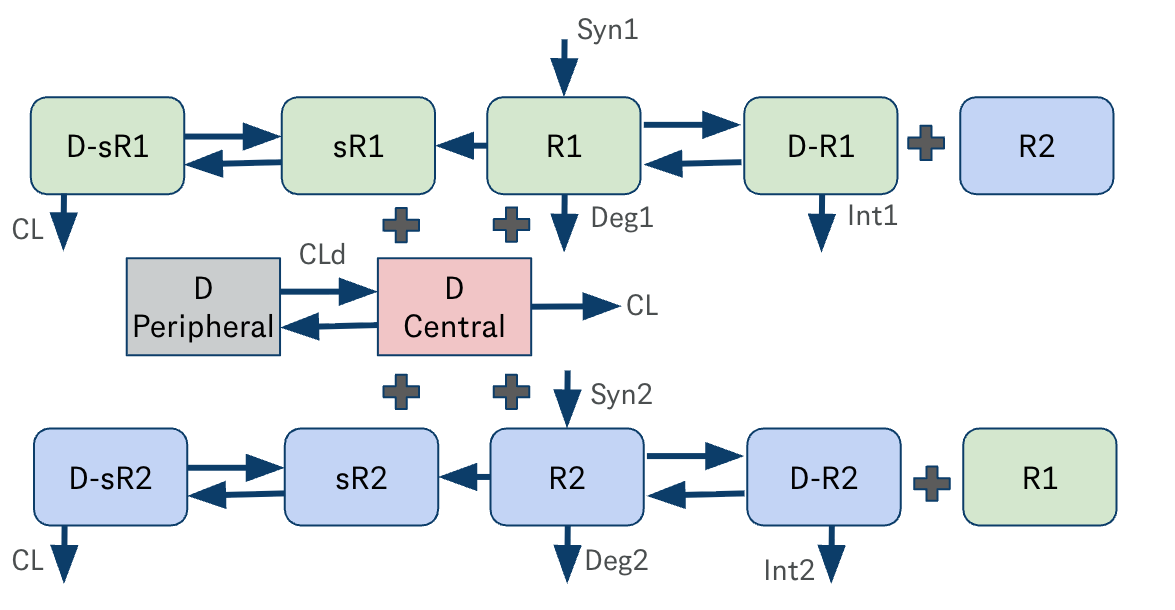}
  \caption{Model expansion after Prompt 2 (C.3): Integration of dual-target TMDD system with R2 receptor. This figure illustrates the system's ability to handle complex multi-target pharmacology with independent receptor dynamics, competitive drug binding, and parallel TMDD pathways while maintaining mathematical consistency across all biological processes.}
  \label{fig:ex3-tmdd-r2}
\end{figure}

\begin{figure}[h]
  \centering
  \includegraphics[width=0.8\linewidth]{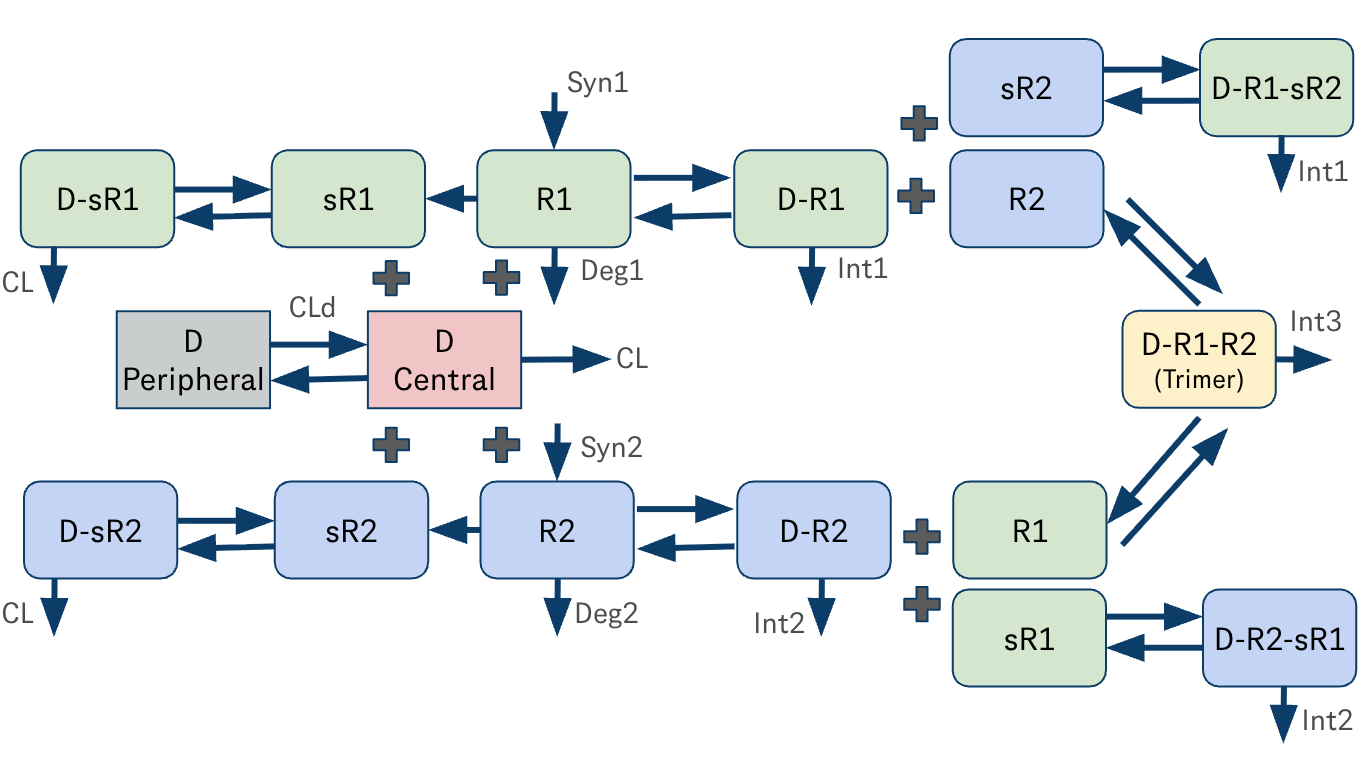}
  \caption{Final model configuration after Prompt 3 (C.4): Implementation of cooperative trimer formation mechanisms. This comprehensive model demonstrates GRASP's capability to handle the most sophisticated biological scenario, incorporating sequential binding processes, cooperative kinetics, and complex stoichiometric relationships while preserving all previous model components and maintaining biological constraint satisfaction.}
  \label{fig:ex4-trimer}
\end{figure}

The progressive model evolution demonstrates GRASP's systematic approach to biological complexity management, where each conversational interaction builds upon the previous model state while preserving existing biological relationships and maintaining mathematical consistency. The visual progression from simple two-compartment pharmacokinetics to complex cooperative binding mechanisms illustrates the framework's capability to handle sophisticated pharmacological modeling through natural language interactions.

\begin{figure}[t]
  \centering

  \begin{subfigure}[b]{0.48\textwidth}
    \centering
    \includegraphics[width=\textwidth]{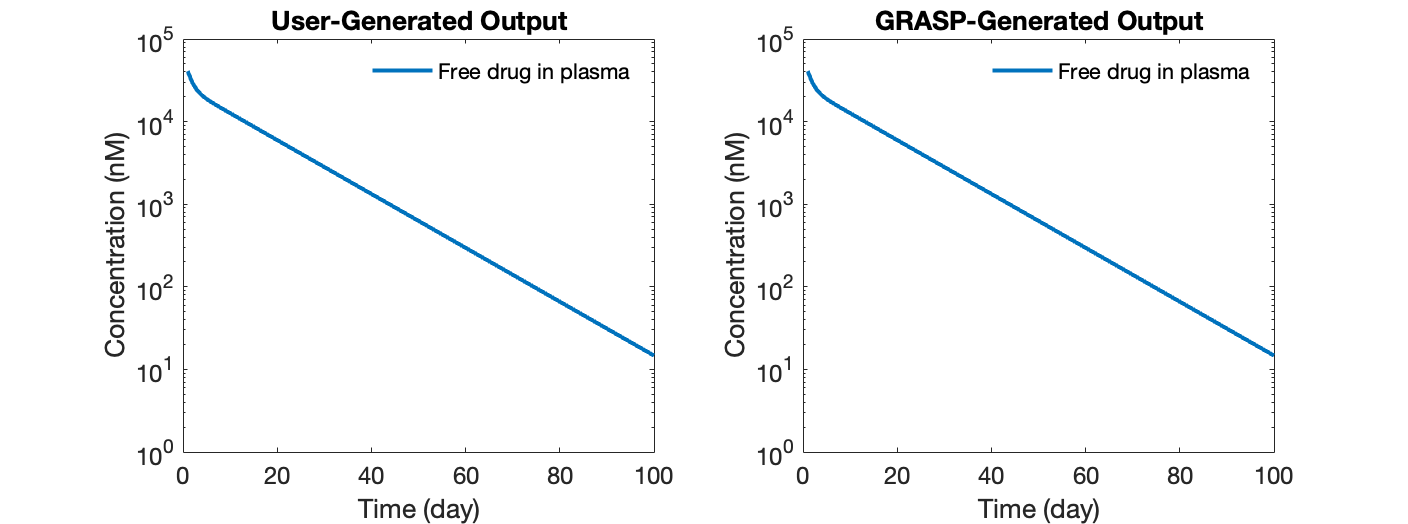}
    \caption{Base two-compartment model}
    \label{fig:base-model}
  \end{subfigure}
  \hfill
  \begin{subfigure}[b]{0.48\textwidth}
    \centering
    \includegraphics[width=\textwidth]{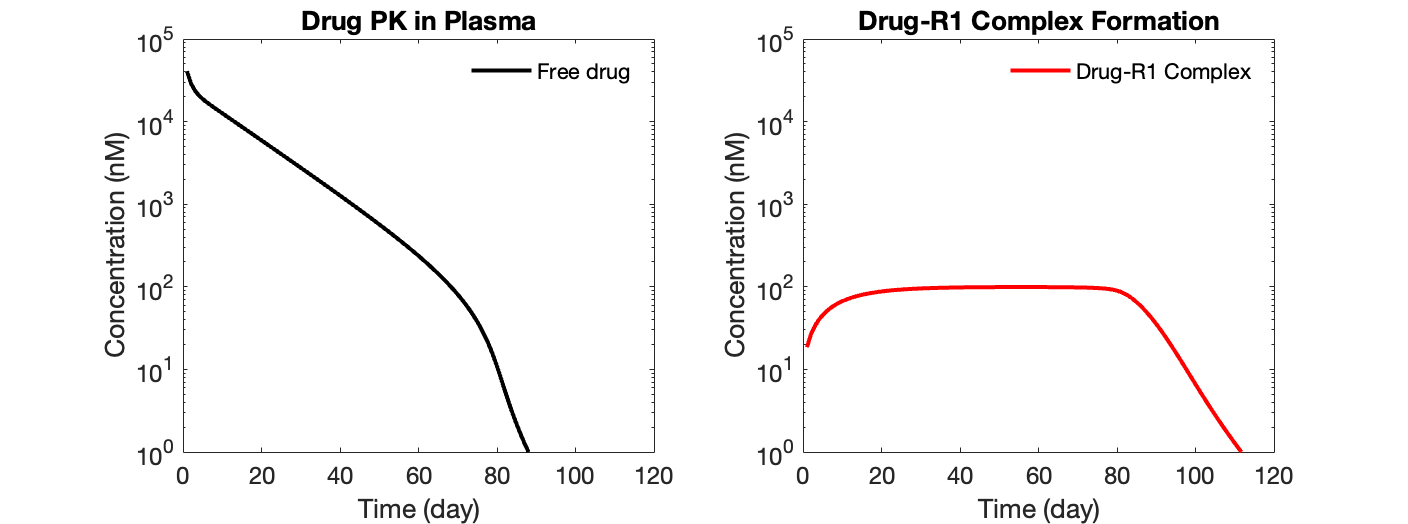}
    \caption{TMDD with R1 receptor (R1)}
    \label{fig:tmdd-r1}
  \end{subfigure}

  \vspace{0.5cm}

  \begin{subfigure}[b]{0.48\textwidth}
    \centering
    \includegraphics[width=\textwidth]{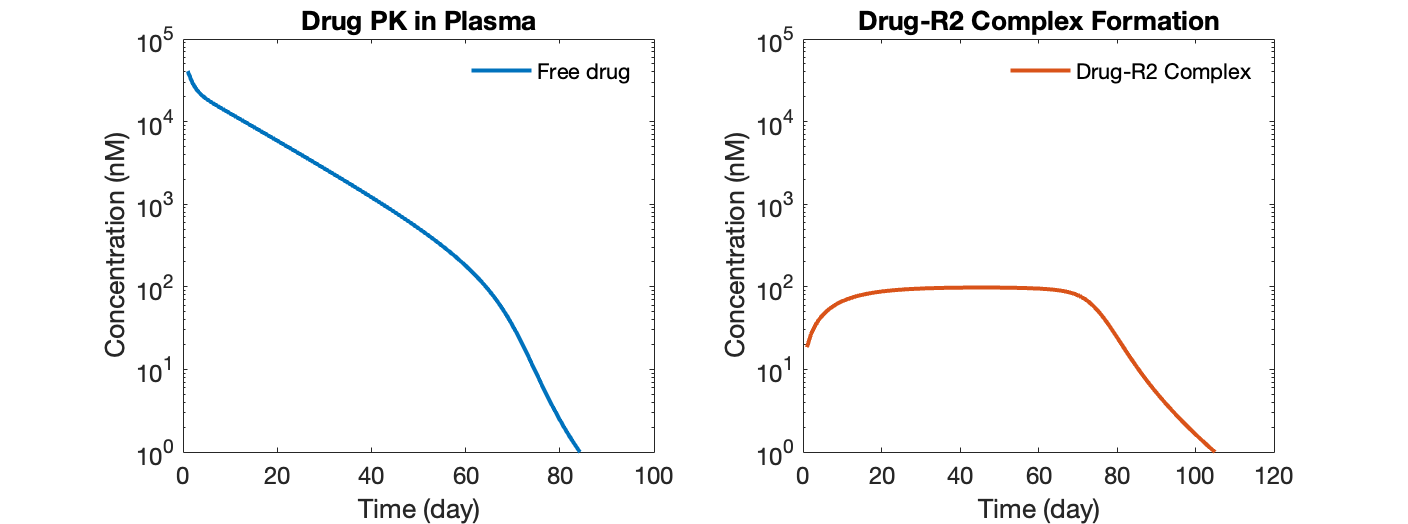}
    \caption{Dual-target TMDD (R1 + R2)}
    \label{fig:tmdd-r2}
  \end{subfigure}
  \hfill
  \begin{subfigure}[b]{0.48\textwidth}
    \centering
    \includegraphics[width=\textwidth]{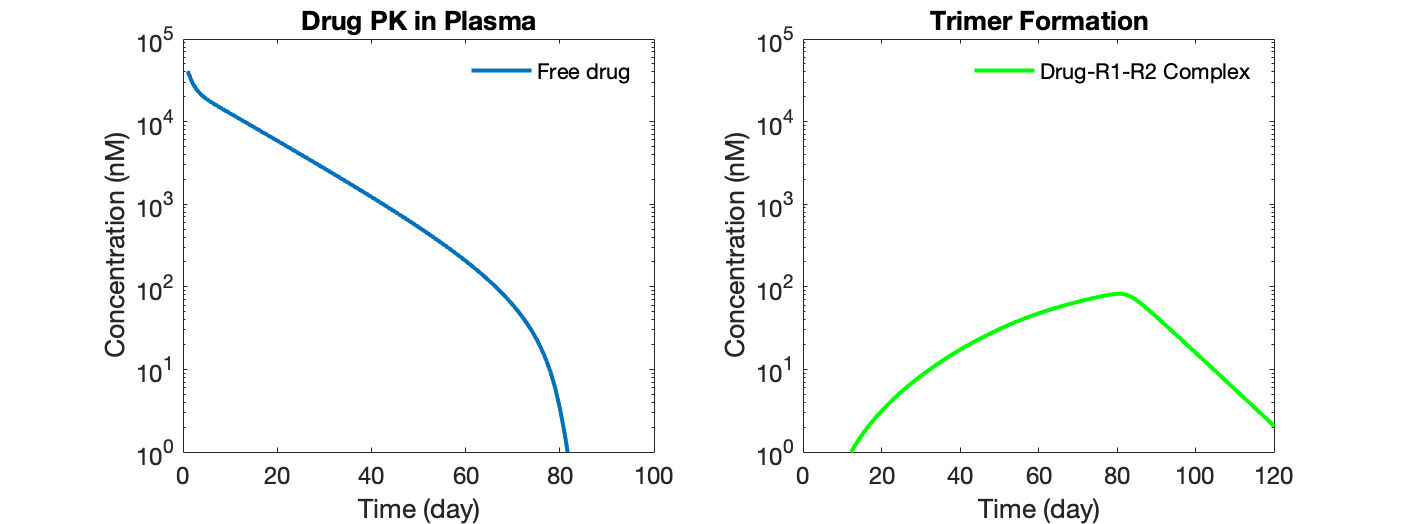}
    \caption{Cooperative trimer formation}
    \label{fig:trimer}
  \end{subfigure}

  \caption{Progressive complexity case study demonstrating GRASP's natural language-driven model modification capabilities. (a) Base two-compartment pharmacokinetic model showing equivalence between user-generated and GRASP-generated outputs. (b) TMDD model with R1 binding (KD = 1 nM) after Prompt 1, demonstrating nonlinear pharmacokinetics fidelity to theoretical TMDD behavior. (c) Dual-target TMDD model after Prompt 2, incorporating both R1 and R2 (KD = 10 nM) binding with competitive kinetics and mass balance preservation. (d) Complete model with cooperative trimer formation after Prompt 3, representing the most sophisticated biological scenario with multi-step assembly processes and complex stoichiometric relationships.}
  \label{fig:simulation-results}
\end{figure}

\clearpage

\section*{Appendix D: Mathematical Foundation: Graph Representation Theory}

We establish the theoretical foundation for graph-based QSP modeling through analysis of representation completeness, biological constraint preservation, and computational complexity.

\textbf{Definition 1 (Extended QSP Model).}
A QSP model $M$ is a 7-tuple $M = (C, S, P, R, \mathcal{F}, \mathcal{B}, \Sigma)$ where:
\begin{itemize}
\item $C = \{c_1,\dots,c_n\}$ is the set of compartments with volumes $V_c(t) \in \mathbb{R}_{>0}$;
\item $S = \{s_1,\dots,s_m\}$ is the set of species with amounts (or concentrations) $x_i(t) \in \mathbb{R}_{\ge 0}$;
\item $P = \{p_1,\dots,p_k\}$ is the set of parameters with values $p_i \in \mathbb{R}_{\ge 0} \cup \mathcal{D}$ (where $\mathcal{D}$ are distributions);
\item $R = \{r_1,\dots,r_\ell\}$ is the set of reactions with rates $v_j = f_j(x,P,t)$;
\item $\mathcal{F}$ is the set of kinetic function types (mass-action, Michaelis–Menten, Hill, etc.);
\item $\mathcal{B}$ is the set of biological constraints (mass balance, thermodynamics, stoichiometry, unit consistency);
\item $\Sigma$ is a symbol table linking kinetic templates to $(S,P,C)$.
\end{itemize}

\textbf{Definition 2 (QSP Hypergraph).}
A QSP hypergraph is $\mathcal{H} = (V, \mathcal{E}, w, \Phi, \Psi)$ with:
\begin{itemize}
\item $V = V_C \cup V_S \cup V_P \cup V_R \cup V_F$ (compartments, species, parameters, reactions, function-type vertices);
\item $\mathcal{E} \subseteq 2^V$ the hyperedge set encoding multi-entity relations;
\item $w:\mathcal{E}\!\to\!\mathbb{Z}$ edge attributes (e.g., stoichiometric coefficients; $w(e)>0$ for products, $w(e)<0$ for reactants);
\item $\Phi: V \!\to\! \mathcal{A}\times\mathcal{T}$ vertex attributes (values, units, and optional time-dependence);
\item $\Psi: \mathcal{E}_R \!\to\! \mathcal{F}$ maps reaction-incident hyperedges to kinetic templates (or equivalently to $V_F$).
\end{itemize}

\textbf{Theorem 1 (Graph Representation and Validation).}
For any well-formed QSP model $M = (C,S,P,R,\mathcal{F},\mathcal{B},\Sigma)$ satisfying $\mathcal{B}$, there exists a QSP hypergraph $\mathcal{H}$ such that:
(i) $M$ can be reconstructed from $\mathcal{H}$ (lossless encoding under $\mathcal{F}$ and $\Sigma$);
(ii) mass-balance feasibility is preserved (existence of a nonnegative mass vector consistent with internal reactions);
(iii) an iterative local-repair process that is monotone and inflationary with respect to $\mathcal{B}$ converges to a fixed point; under a geometric decrease assumption it achieves a logarithmic iteration bound.

\textbf{Proof.}
Construct a mapping $\mathcal{T}: M \rightarrow \mathcal{H}$ in four steps.

\emph{Step 1 (Vertex construction).}
\begin{align}
V_C &= \{v_c : c\!\in\!C\}, \ \ \Phi(v_c)=(V_c(t),\text{compartment}); \\
V_S &= \{v_{s_i}: s_i\!\in\!S\}, \ \ \Phi(v_{s_i})=(x_i(0),\text{species},\text{units}); \\
V_P &= \{v_{p_i}: p_i\!\in\!P\}, \ \ \Phi(v_{p_i})=(p_i,\text{parameter},\text{units}); \\
V_R &= \{v_{r_j}: r_j\!\in\!R\}, \ \ \Phi(v_{r_j})=(\text{rate symbol }v_j,\text{reaction}); \\
V_F &= \{v_{f}: f\!\in\!\mathcal{F}\}, \ \ \Phi(v_f)=(\text{kinetic template}).
\end{align}
Associate each $v_{r_j}$ with a template via $\Psi$ (or an incident edge to $v_f$).

\emph{Step 2 (Hyperedge construction).}
For each reaction $r_j$, create reactant/product hyperedges connecting $v_{r_j}$ to involved $v_{s_i}$ and assign $w(e)=S_{ij}$, the integer stoichiometric coefficient (negative for reactants, positive for products). Transport and compartmental flows are encoded by hyperedges linking species across $V_C$ with attributes capturing flow rates and volume dependence. This induces a sparse stoichiometric matrix $S \in \mathbb{Z}^{m\times \ell}$ recoverable from $\{w(e)\}$.

\emph{Step 3 (Mass-balance validation).}
Let $\mu \in \mathbb{R}^m_{\ge 0}$ be molecular masses (per species). Internal reactions conserve mass iff $\mu^\top S = 0$. With inter-compartmental transport and time-varying volumes $V_c(t)$, total mass changes only by explicit source/sink and flow terms; the check reduces to a pass over $\mathrm{nnz}(S)$ and associated transport edges. Thus validation is $O(\mathrm{nnz}(S))$.

\emph{Step 4 (Iterative convergence).}
Define a nonnegative violation functional $\varepsilon(\mathcal{H})$ as the sum of residuals of local predicates in $\mathcal{B}$ (mass balance, unit/thermo checks, connectivity). Let $\mathcal{K}$ be a local repair operator that propagates updates along incident hyperedges and is monotone and inflationary (each application does not retract satisfied predicates and does not increase $\varepsilon$).
Then repeated application of $\mathcal{K}$ converges to a fixed point (no further violations) in finitely many steps because the attribute lattice is finite on a fixed $\mathcal{H}$. If, in addition, $\mathcal{K}$ ensures a geometric decrease $\varepsilon_{t+1}\!\le\!\rho\,\varepsilon_t$ for some $\rho\!\in\!(0,1)$ (Assumption A1, commonly met by halving-style unit/stoichiometry adjustments), then the number of iterations to reach $\varepsilon\!\le\!\varepsilon_{\text{target}}$ is bounded by
\[
K \le \left\lceil \log_{1/\rho}\!\left(\frac{\varepsilon_0}{\varepsilon_{\text{target}}}\right) \right\rceil.
\]

\textbf{Corollary 1 (Biological constraint preservation).}
Any transformation $T:\mathcal{H}\!\to\!\mathcal{H}'$ that preserves (a) hyperedge incidence, (b) stoichiometric weights $w(e)$, (c) kinetic template labels $\Psi$, and (d) units and compartment labels in $\Phi$ maintains validity with respect to $\mathcal{B}$; i.e., $\mu^\top S=0$ remains feasible and unit/thermo predicates remain satisfied.

\textbf{Corollary 2 (Scalability).}
Assuming sparsity, representation and validation scale as
memory $O(|V| + |\mathcal{E}| + \mathrm{attr})$ and time $O(\mathrm{nnz}(S) + |\mathcal{E}_{\text{transport}}|)$.
For large but sparse models (e.g., up to $10^6$ species and $10^5$ reactions with bounded average degree and compact attribute payloads), the induced $\mathrm{nnz}(S)$ dominates complexity, yielding near-linear passes in practice.

\medskip
This foundation yields a lossless graph encoding of QSP models under a fixed kinetic grammar, a correct mass-balance criterion recoverable from the hypergraph, and a convergent local-repair process with a logarithmic iteration bound under a mild geometric-decrease assumption.


\end{document}